\documentclass[conference]{IEEEtran}

\usepackage[pdftex]{graphicx}
\usepackage{amsmath}
\usepackage{eqparbox}
\usepackage{booktabs}
\usepackage{algorithm}
\usepackage[noend]{algpseudocode}
\usepackage{fancyhdr}
\fancypagestyle{specialfooter}{%
  \fancyhf{}
  
  \lfoot{\fontsize{8}{8} \selectfont © 2020 IEEE. Personal use of this material is permitted. Permission from IEEE must be obtained for all other uses, in any current or future media, including reprinting/republishing this material for advertising or promotional purposes, creating new collective works, for resale or redistribution to servers or lists, or reuse of any copyrighted component of this work in other works. }
}

\begin{document}


\title{\LARGE FPGA Implementation of Simplified Spiking Neural Network}

\author{\IEEEauthorblockN{Shikhar Gupta, Indian Institute of Technology, Guwahati\\ Arpan Vyas, Indian Institute of Technology, Guwahati \\ Gaurav Trivedi, Indian Institute of Technology, Guwahati}
\IEEEauthorblockA{Email: guptashikhar20@gmail.com, arpanvyas@gmail.com, trivedi@iitg.ac.in}}


\maketitle



\begin{abstract}
Spiking Neural Networks (SNN) are third-generation Artificial Neural Networks (ANN) which are close to the biological neural system. In recent years SNN has become popular in the area of robotics and embedded applications, therefore, it has become imperative to explore its real-time and energy-efficient implementations. SNNs are more powerful than their predecessors because they encode temporal information and use biologically plausible plasticity rules. In this paper, a simpler and computationally efficient SNN model using FPGA architecture is described. The proposed model is validated on a Xilinx Virtex 6 FPGA and analyzes a fully connected network which consists of 800 neurons and 12,544 synapses in real-time.


\end{abstract}


\section{Introduction}
\thispagestyle{specialfooter}

With Moore's law being transformed into more than Moore, the scientific community is exploring alternate avenues for faster, cheaper, and more efficient computing, and Neuromorphic Computing is found to be one of the viable replacements of the present computing paradigm. Neuromorphic computing can be realized efficiently by utilizing Spiking Neural Networks (SNN). It aims at realizing the architecture and performance of a brain in silicon. The brain, being the most efficient system present, processes complex information much faster than any existing computer. In recent years, the popularity and application of spiking neural networks have increased considerably. SNN is prominently used in many applications such as event detection, classification, speech recognition, spatial navigation, and autonomous motor control. It has demonstrated its effectiveness in detecting analog signals from sensors \cite{1}; designing controllers for autonomous robots \cite{3}; performing detection and recognition tasks \cite{5}; processing cortical data \cite{7} and tactile form-based recognition \cite{8}.

With the advent of autonomous robots and self-driving vehicles and due to the rise in the realtime applications of embedded systems, it has become imperative to realize machine learning models on compact and energy-efficient platforms. The existing neural network models \cite{9}, are computationally intensive and require huge memory for their realization, therefore making them unsuitable for realtime and energy-efficient applications. Although SNN has been realized in Applications Specific Integrated Circuits (ASIC) such as SpiNNaker \cite{10}, BrainScaleS \cite{11}, SyNAPSE \cite{12} Neuropipe-chip \cite{13}, etc. their objective is mainly to provide a solution for large scale simulations rather than for low power embedded applications. The SNN model proposed in this paper is aimed towards its energy-efficient, portable and realtime implementation for improvising the performance of electronic systems. 

Conventionally, Field Programmable Gate Arrays (FPGA) are utilized for the validation of electronic systems as well as in the implementation of time-critical systems. FPGAs are also well-suited for providing a low power solution for massively parallel and computationally less complex models \cite{omondi2006fpga}. Although parallelism can be achieved using Graphics Processing Units (GPU) as well, FPGA implementations are advantageous where power consumption is an issue \cite{power}. 

In today's era, it is becoming mandatory for the intelligent system to learn efficiently in realtime from its surroundings. Existing FPGA models of SNN \cite{15,16} do not support the training of the network and, if the weights must be changed, the hardware has to be reprogrammed, rendering it unsuitable for realtime applications. The proposed solution combines the neuron membrane model and on-line spike-time dependent plasticity (STDP) learning.

\begin{table*}[h!]
\begin{center}
    \caption{Results of relevant state-of-the-art SNN hardware accelerators}
    \label{tab:time_res}
    \begin{tabular}{ l | l | l | l | l | l | l }
    Work & Platform & Model & Time Resolution & Neurons & Synapses/Neuron & Resources/Neuron (FF, LUT) \\
    \toprule
    Upegui et al., 2005 & FPGA & Custom & $ 1ms $ & 30 & 30 & 100, 100 \\
    Pearson et al., 2007 & FPGA & LIF & $ 0.5ms $ & 1120 & 912/112 & -- \\
    Cassidy et al., 2007 & FPGA & LIF & $ 320ns $ & 51 & 128 & 146, 230 \\
    Jin et al., 2008 & Multiprocessor (ARM) & IZ & $ 1ms $ & 1000 & 100 & -- \\
    Thomas and Luk, 2009 & FPGA & IZ & $ 10us $ & 1024 & 1024 & 39, 19 \\
    Ambroise et al., 2013 & FPGA & IZ & $ 1ms $ & 117 & 117 & 8, 17 \\
    Cheung et al., 2016 & FPGA & IZ & $ 1ms $ & $>$ 98,000 & 1,000 - 10,000 & -- \\
    Pani et al., 2017 & FPGA & IZ & $ 0.1ms $ & 1440 & 1440 & 37, 39 \\
    This work & FPGA & Simplified LIF & $ 1ms $ & 800 & 12,544 & 29, 70 \\ 
    \bottomrule
    \end{tabular}
    
\end{center}
\end{table*}

\section{Related Work}
Multiple hardware accelerators for SNN have been designed and implemented. The central aim of these architectures is to eradicate the limitations of software simulators like BRIAN \cite{20}, NEST \cite{21}, and NEURON \cite{22}, which are widely accepted in the research community. However, they hit scaling issues and become too slow for large scale networks due to lack of parallel computations \cite{23}. In this section, we present an account of the previously implemented architectures. Table \ref{tab:time_res} summarizes the similar works and corresponding time resolution, number of neurons, and synapses simulated and resources consumed per neuron. Comparing resources used per neurons takes into account the size of the network and is favorable for bigger networks that consume more resources. 

The FPGA architecture by \cite{27} simulates a fully connected network of 1,440 Izhikevich (IZ) neurons. The design updates all neurons at every time step regardless of the spiking activity. The time resolution recorded is 0.1 ms. Our proposed design is event-driven and updates a neuron only if there is any spiking activity. This implementation is more energy-efficient and gives two orders of magnitude lesser time resolution (2.5 us). In terms of resource usage, our architecture (800 neurons, 12,544 synapses) is far more efficient than theirs (1,440 neurons, 16 synapses). Also, their fixed-point representation of weights provides lesser flexibility than our floating-point representation. Although floating-point has higher latency, simplified equations compensated for it.

Another implementation is NeuroFlow \cite{24}, which can simulate an impressive number of neuronal units (600,000), both LIF and IZ, on a 6-FPGA system. When compared to this setup, our design provides a sampling rate of 400kHz against their 1kHz. 

The same applies to large scale hardware implementations like SpiNNaker \cite{10}, which are not well suited for low-power, compact embedded applications and cost very much. Our implementation aims at providing a comparatively small scale and energy-efficient solution. On the other hand, several neural implementations based on Application Specific Integrated Circuits (ASIC) have also been proposed \cite{25}. Although they exhibit better performance and are more energy-efficient, there is a growing interest in the use of FPGA for this. FPGA provides the user the freedom to reconfigure fully or partially the configuration bitstream. Also, they facilitate the creation of multiprocessor systems (as on ASIC) due to the presence of IP cores \cite{27}.


\section{Simplified Neural Network Model}

As far as classic leaky integrate-and-fire (LIF) neuron model \cite{9} is concerned, it is computationally very complex and has large memory requirements. In the literature \cite{18}, is described a simplified version of the LIF model with computationally less complex membrane potential update equations. Let $P_{t}$ be the membrane potential (function of time) which is altered by each incoming spike $S_{it}$, $i=[1...n]$ by a value of synapse weight $W_{i}$, along with the voltage leakage factor, $D$,
\begin{eqnarray}
\begin{footnotesize}
P_{t} = 
\begin{cases}
    P_{t-1} + \displaystyle\sum_{i=1}^{n} S_{it}W_{i} - D     \quad \text{if } P_{min} > P_{t-1} < P_{threshold}\\
    P_{refract}   \quad \quad \quad \quad \quad \quad \quad  \text{if } P_{t-1} \geq P_{threshold}\\
    R_{P}   \quad \quad \quad \quad \quad \quad \quad \quad \quad \text{if } P_{t-1} \leq P_{min}
\end{cases}
\end{footnotesize}
\end{eqnarray}
At every time constant $t$, the membrane potential decreases by a fixed factor $P_{t}$ = $P_{t-1}$ - $D$ given that $P_{t-1}$ is greater than $R_{p}$, the resting potential. This simplified equation can be easily implemented in hardware  unlike classic models which require large number of look up tables. When $P_{t}$ $>$ $P_{threshold}$ a spike is fired and the neuron transits into refractory phase where it blocks any input for a duration of $t_{refract}$.


\section{Spike Time Dependent Plasticity}
Spike Time Dependent Plasticity (STDP) is a biological process, first discovered by Bi and Poo \cite{19}, which changes the connection strengths between neurons (synapses) in the brain. It is an unsupervised learning algorithm that operates upon the time difference between post-synaptic and pre-synaptic spikes. In a given synapse, if the post-synaptic spike occurs in a specific time window (STDP window) after pre-synaptic spike, the synaptic strength is increased, and if it occurs before pre-synaptic spike, the strength is decreased. 

Equation 2 describes the simplified weight change rule. STDP window is described as $ t  \in  [2, 20] $ in both directions. Weights are kept within the range $ w_{min} < w < w_{max} $.
\begin{eqnarray}
w_{new} = 
\begin{cases}
    w_{old} + \sigma \Delta w(w_{max} - w_{old})     \quad \text{if } \Delta w > 0\\
    w_{old} + \sigma \Delta w(w_{old} - w_{min})     \quad \text{if } \Delta w \leq 0
\end{cases}
\end{eqnarray}
$\Delta w$ is calculated using the exponential reinforcement curve employed in classic LIF. It has only 19 entries on either side and, hence, requires very few memory resources (lookup tables) for implementation. Equation 3 describes the determination of change in weights; $\Delta t$ is the time difference between pre-synaptic and post-synaptic spikes and $A^{+}$ and $A^{-}$ are the constants for positive and negative $\Delta t$ values, respectively; $\tau_{+}$ and $\tau_{-}$ are steepness time constants in both directions. Weights are kept within the range $w_{min} < w < w_{max}$.
\begin{eqnarray}
\Delta w = 
\begin{cases}
    A^{-} \exp(\frac{\Delta t}{\tau_{-}})     \quad \text{if } \Delta t \leq -2\\
    0  \hspace{0.99cm} \quad \quad \quad \text{if } 2 <\Delta t < 2\\
    A^{+} \exp(\frac{\Delta t}{\tau_{+}})     \quad \text{if } \Delta t \geq -2
\end{cases}
\end{eqnarray}



\section{Key Design Elements}

Spike Time Dependent Plasticity is the backbone of the spiking neural networks as it enables the learning process. A complete system aiming to replicate the biological neural model requires feature extraction, relativity of neural activity based on input strength, and competition among neurons for a particular class. These are described as follows.

\subsection{Visual Receptive Field}
Receptive Field is the area of an image which produces the input for a visual neuron. As the neural layers go deeper the receptive fields get bigger, convolving inputs from the previous layers. It can be expressed as a convolution filter for a variety of operations like edge detection, sharpening, and blurring. In our system we have implemented the receptive field, RF as a low pass blurring filter on the image.

\begin{figure}[ht!] 
\includegraphics[width=3.4in]{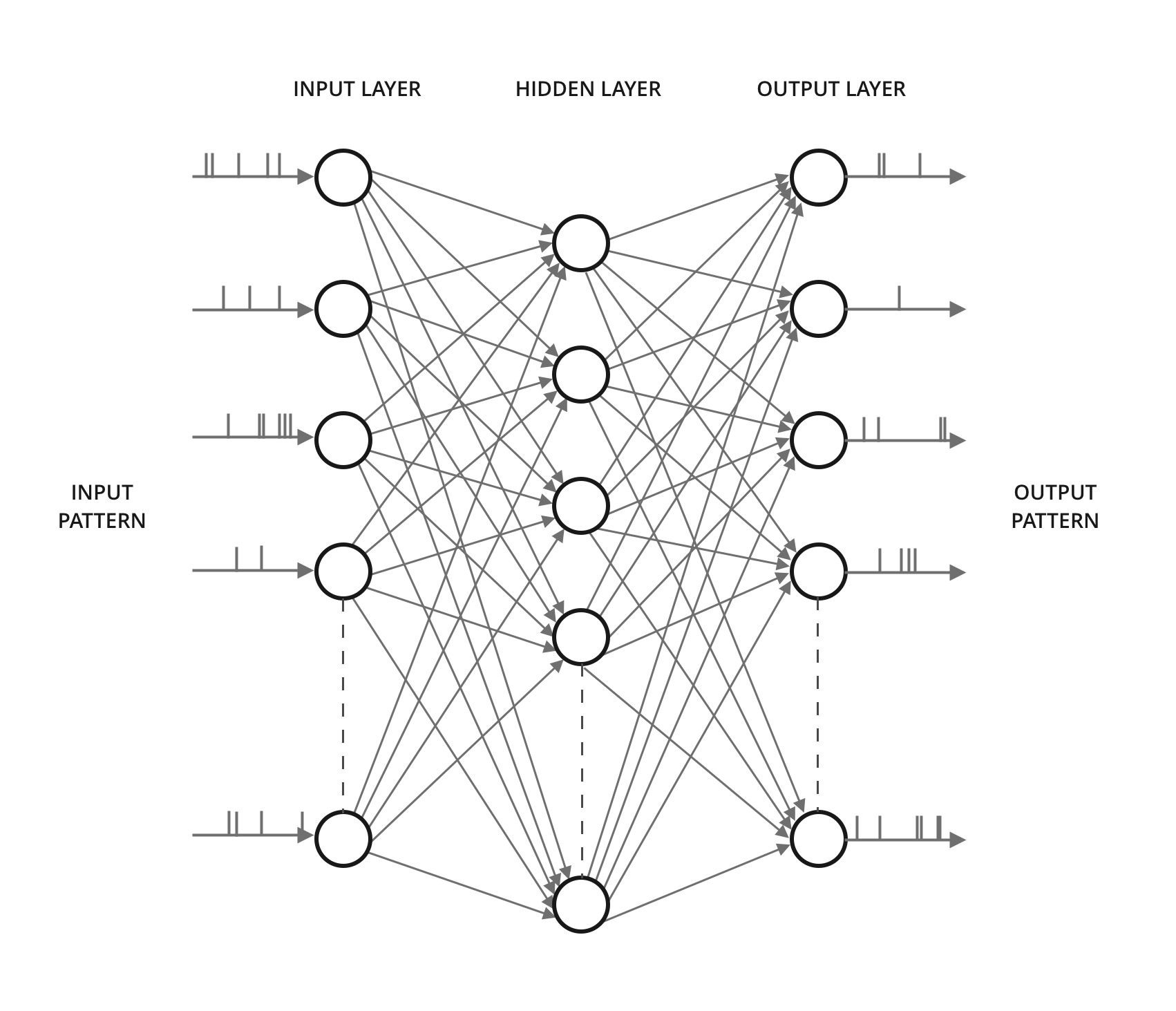}

\caption{Full Mesh Spiking Neural Network}
\label{fig:full_mesh}
\end{figure}

\subsection{Spike Generation}
In a biological neuron, the input excitation is transmitted as time-domain impulses. The frequency of the impulses is proportional to the excitation. We have replicated this by taking the output of the receptive field $RF$, the maximum membrane potential $R_{max}$, and the minimum refractory period $RP_{min}$ to generate firing rate of the neuron $FR$.
\begin{eqnarray}
FR = 
\begin{cases}
    \frac{1}{RP_{min}*\frac{RF}{R_{max}}}   \quad \text{if }  RF > 0\\
    0		 \quad  \quad  \quad   \quad  \quad \quad \text{if }  RF \leq 0
\end{cases}
\end{eqnarray}


\subsection{Variable Threshold and Lateral Inhibition}
The voltage threshold at which a neuron fire is kept variable for each image. This is to make sure that images irrespective of their relative brightness produce uniform output for subsequent layers. The voltage threshold is kept as one third the maximum input spike frequency of the layer.

Lateral Inhibition is used to induce competition among output layer neurons for a particular class. When the first output neuron fires for a particular image, the potential of all the other neurons is reduced by half the threshold potential. This 'winner takes all' strategy ensures selectivity of the winner neuron for the class as the firing activity of other contending neurons is suppressed. Lateral inhibition is from biological neural networks where strengthening of one neural pathway weakens the neighbouring ones.

\subsection{Hardware Algorithm}
\begin{figure}[ht!] 
\raggedleft
\includegraphics[width=3.6in]{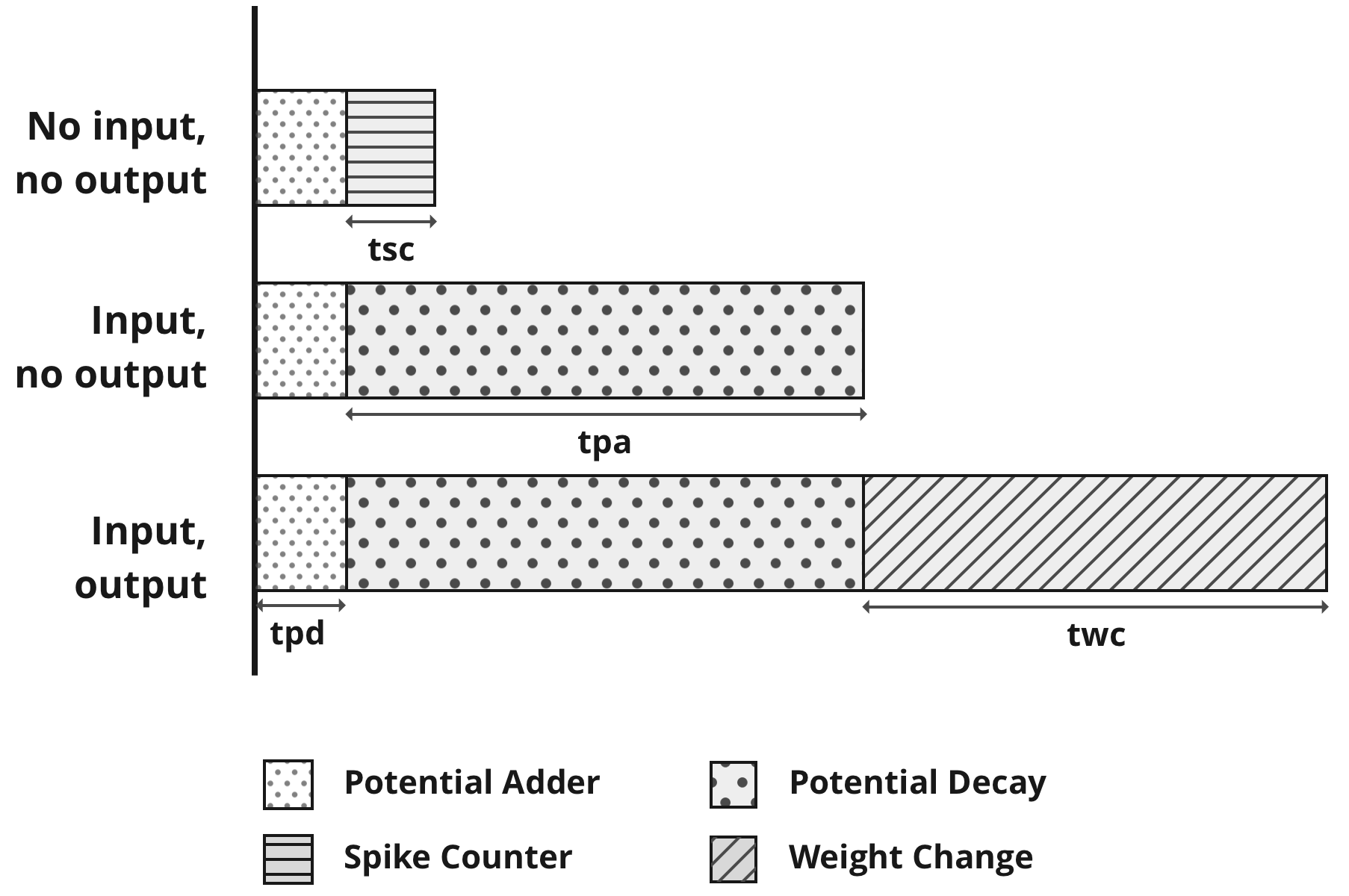}

\caption{Breakdown of a Time Unit into processes under different scenarios. X axis denotes time and y axis points to the scenarios. Clock cycles consumed by a Time Unit differ depending upon input and output, being the least when there is neither an input nor an output spike and the most when both input and output spikes are present. \textit{tsc}, \textit{tpa}, \textit{tpd} and \textit{twc} are the actual times taken by each process. The figure is according to scale i.e \textit{tsc} = \textit{tpd} \textless \textit{tpa} \textless \textit{twc}}
\label{fig:time_units}
\end{figure}

Spiking neural networks rely on the temporal information carried by spike trains and hence require a notion of time. A Time Unit block has been used in our system to keep track of time steps during training and classification. Although the number of time units is fixed for every data sample in training or classification phases, the number clocks to process one is highly adaptive as shown in Fig.\ref{fig:time_units}. 


\begin{figure}[ht!] 
\centering
\includegraphics[width=3.4in]{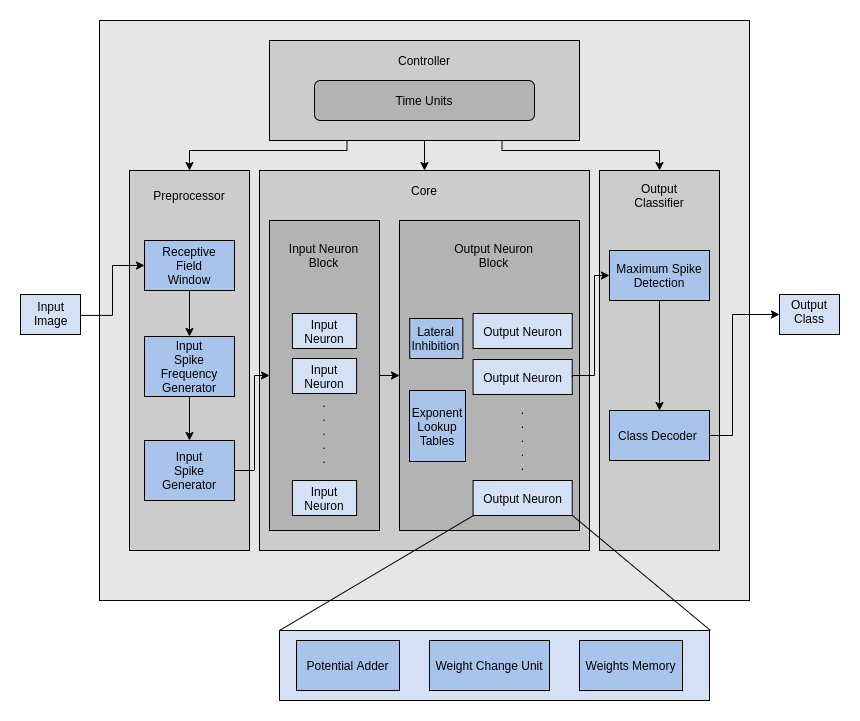}

\caption{Block Level view of the system}
\label{snn_block}
\end{figure}


\section{Results}

\subsection{Resource Usage}
The network with 784 input ($I$) and 16 output ($O$) neurons and with 12,544 synapses using 24 bits to represent a membrane potential and weights ($W$) was implemented on Xilinx XC6VLX240T device. A single Block RAM of size 36Kb was used as FIFO for each output neuron to store synapse weights. Additionally, four DSP48E1 slices capable of addition and multiplication were used for single output neuron in its Potential Adder and Weight change blocks. The Table \ref{tab:tableR} provides resource usage for the complete system.
\begin{table}[h!]
\begin{center}
    \caption{Resource Usage}
    \label{tab:tableR}
    \begin{tabular}{ l | l | l | l | l}
    Resource & Used & Available & \% Used & Resource Complexity \\
    \toprule
    Flip-flops & 23,238 & 301,440  & 8\% & $I . O . W $ \\
    Slice LUTs & 56,230 & 150,720 & 37\% & $I . O . W $\\
    BRAMs & 16 & 416 & 4\%& $ O $\\
    DSP48E1 & 64 & 768 & 8\% & $ O$\\
    \bottomrule
    \end{tabular}
    
\end{center}
\end{table}

\subsection{Timing Analysis}
The proposed architecture has exploited the sparsity of spiking neural network for improvement in speed by morphing Time Units dynamic according to the spiking activity in input and output neurons (Fig.\ref{fig:time_units}). Table \ref{tab:table1} and Table \ref{tab:table2} provides timing analysis for classification and training of $N$ images at an operating frequency of 100MHz.

\begin{table}[h!]
\begin{center}
    \caption{Timing Analysis for Classification}
    \label{tab:table1}
    \begin{tabular}{ l | l | l }
    Operation & Time & Time Complexity\\
    \toprule
    Time Unit - Maximum & 8.5us & $ I.O.W $ \\
    Time Unit - Average & 2.5us & $ I.O.W$ \\
    Single Image & 0.5ms & $I.O.W$ \\
    Classifying 10,000 images & 5s & $ N.(I.O.W) $\\
    \bottomrule
    \end{tabular}
    
\end{center}
\end{table}

\begin{table}[h!]
\begin{center}
    \caption{Timing Analysis for Training}
    \label{tab:table2}
    \begin{tabular}{ l | l | l }
    Operation & Time & Time Complexity \\
    \toprule
    Time Unit - Maximum & 17us & $ I.O.W $  \\
    Time Unit - Average & 5us & $ I.O.W $ \\
    Single Image & 1.1ms & $I.O.W$ \\
    Training 60,000 images & 65s & $ N.(I.O.W) $ \\
    \bottomrule
    \end{tabular}
    
\end{center}
\end{table}

Table \ref{tab:cpu} presents the comparison of training and classification timings of a single image from the MNIST dataset over FPGA and CPU. Our FPGA architecture showcases a speedup of 256x for classification and 187x for training. The results account for the massive parallelism provided by FPGA at a low energy cost.

\begin{table}[h!]
\begin{center}
    \caption{Comparison with Serial Computation}
    \label{tab:cpu}
    \begin{tabular}{ l | l | l }
    Operation & FPGA & CPU \\
    \toprule
    Classification & 0.5ms & $ 128ms $  \\
    Training & 1.08ms & $ 202ms $ \\
    \bottomrule
    \end{tabular}
    
\end{center}
\end{table}


\section{Conclusion}
In this paper, we described an architecture for Simplified Spiking Neural Network which is implemented on FPGA and optimized for low power embedded applications with real-time learning. The simplification of the STDP algorithm doesn't compromise with the classification and learning capabilities of the network and rather reduces the computation complexity which in turn helps in developing a hardware accelerator with minimum resource usage. The system is designed to take advantage of the sparsity of the network and fabricate each time unit according to the activity in the network. Also, an account of parameter analysis is presented which showcases our methodology of choosing efficient parameter values.

\bibliographystyle{IEEEtran}
\bibliography{biblio_rectifier}
\end{document}